# SIMULATION MODEL OF TWO-ROBOT COOPERATION IN COMMON OPERATING ENVIRONMENT

*V.Ya. Vilisov, B. Yu. Murashkin, A. I. Kulikov*
*University of Technology, Russia, Moscow Region, Korolyov city*
*vvib@yandex.ru*

***Abstract.*** *The article considers a simulation modelling problem related to the chess game process occurring between two three-tier manipulators. The objective of the game construction lies in developing the procedure of effective control of the autonomous manipulator robots located in a common operating environment. The simulation model is a preliminary stage of building a natural complex that would provide cooperation of several manipulator robots within a common operating environment. The article addresses issues of training and research.*

**Introduction.** Modern robotics has a wide range of tasks where robots are to perform various manipulations with objects. These applications include: products' assembly, room cleaning, cargo moving etc. [1, 2].

As a rule, university students of robotics study mostly construction of separate robotic units and their control. Thus, little attention is paid to the issues of robots' cooperation in groups. The main focus of the work is the construction of models and robotic complexes that would allow obtaining necessary research and development skills related to the robot cooperation in groups and their coordinated actions to solve the problems that require participation of several robots and/or heterogeneous "robot-human" groups.

The important subset of such problems is comprised of the operations that are executed by several robots or in robot-human groups. They are also called "collaborative robots" [3, 4]. The main requirement set for such robot-robot groups or robot-human groups is taking into account other group members or coordination of their engagement. Thus, in various assembly operations, robots must adhere to a certain sequence of actions. Therefore, the chess game between two manipulator robots may become a useful model for developing the robot operation algorithms, their sequence of actions etc.. Moreover, using the framework of the model, it is also possible to work over various types of grips and/or to optimize the control system used for grabbing different objects etc. [2]. There are many publications that are devoted to the above-mentioned research [1-4].

Usually, the first stage of solving such problems would include mathematical modelling [4], in particular - simulation modelling. Here we have demonstrated a simulation model of cooperation between two manipulator robots within the chess game environment. The simulation is based on the geometrical dimensions of the robots' tiers, chess game logic and time required to make a move by each robot competitor. Statistical time characteristics pertaining to the moves of each robot may be changed as a part of the simulation model.

Currently, there are several implementation types of mechatronic devices [5-12] that can move chess pieces on the board and choose the move (see Pic. 1). Industrial robots are also used when moving the chess pieces in a robot-human game (see Pic. 1c) or in other games like Go [5] (see Pic. 1d). However, the work is focused on the chess game between two manipulator robots. This problem is a part of a wider research on a robot-robot and robot-human cooperation as well as the robot training.

In order to simulate the chess game between two manipulator robots, we have developed a program that implements and demonstrates the process as a sequence of 3D stages.

**Structure and Combination of Simulation Model.** The program has been developed in *C#* (C Sharp) using ***Unity***, a cross-platform environment for the computer game development [13]. The structure of main program elements is provided on Pic. 2.

Main program file (*Main.cs*) contains settings, interface elements, decoder, separator and program's logic control elements. The decoder transforms chess game text file *Game.txt* (Pic. 3) where the moves are recorded in the way that is customary for users, into the numerical form (file *Game_Symbols.txt*).

The separator splits the *Game_Symbols.txt* file for each manipulator. Thus, *PlayerWhite.txt* file contains all odd array elements (moves of the player with the white chess pieces) that should be executed by the manipulator robot 1 (MR1), while *PlayerBlack.txt* file is created for MR2, containing all even array elements, i.e. the moves of the player with the black chess pieces.

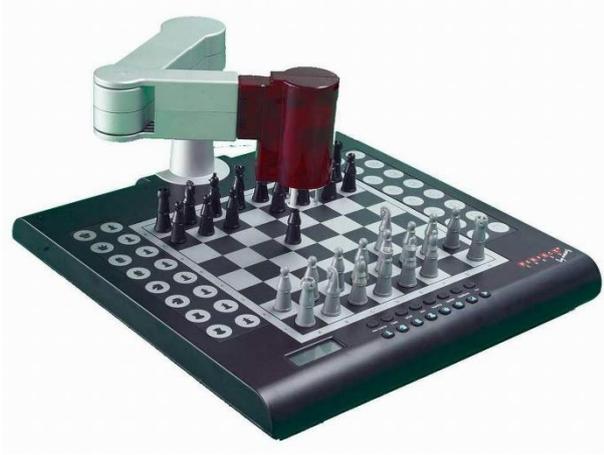

a)

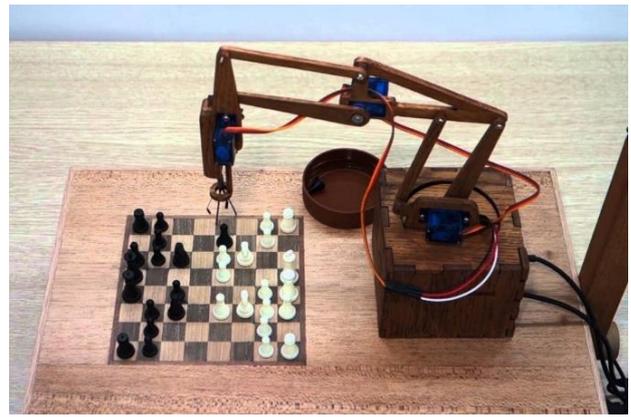

b)

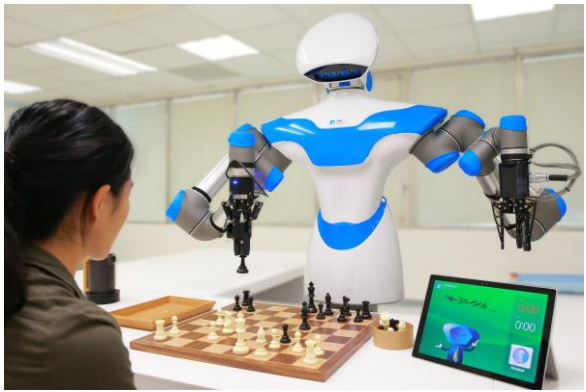

c)

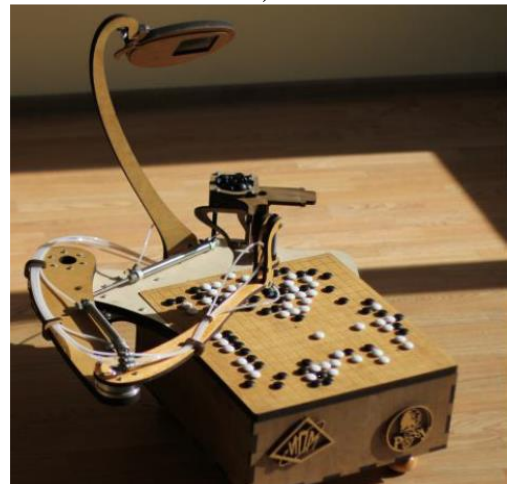

d)

Picture 1 - Mechatronic Players

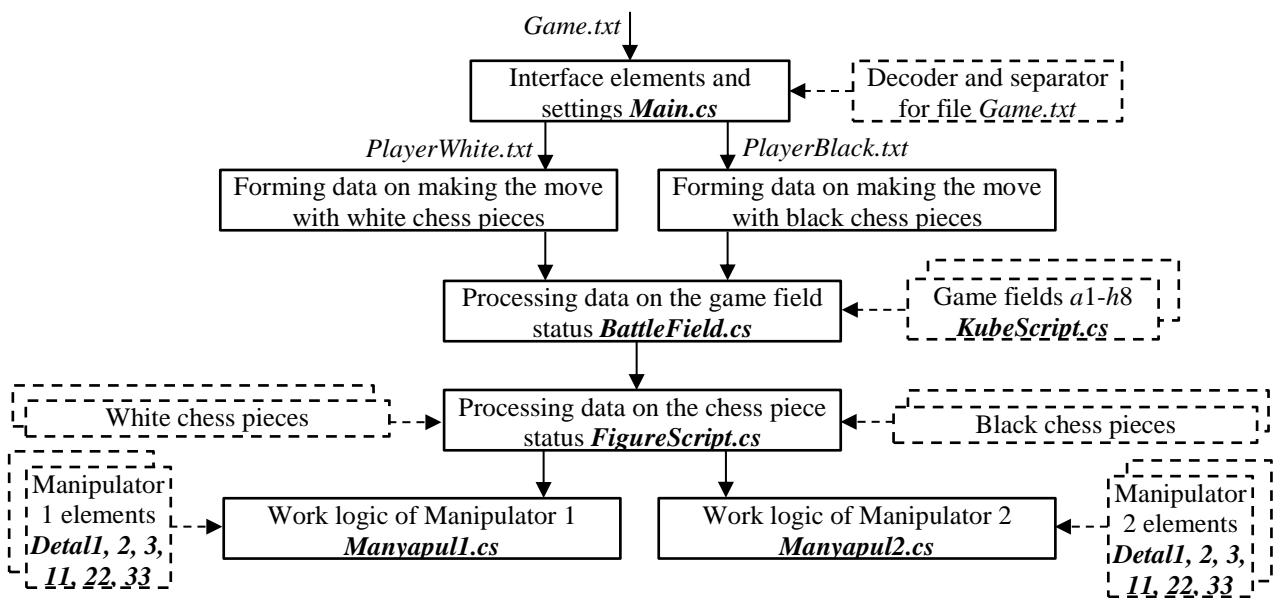

Picture 2 - Simulation Program Object Structure

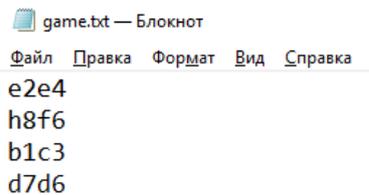

Picture 3 - Part of a Game's Simulation File

Child files of *Main.cs* script, i.e. scripts *PlayerWhite.cs* and *PlayerBlack.cs* read the data from the files *PlayerWhite.txt* and *PlayerBlack.txt* accordingly, receiving the "next move" message from the script *Main.cs* and sending data regarding the players' moves together with the "next move" request to *BattleField.cs* script.

Game object *BattleField* contains script file *BattleField.cs* together with its child game objects, where each of them contains an additional script file *KubeScript.cs* that corresponds to a specific game field square and possesses a square-specific name. Moreover, we have created scripts for the game objects *Manipulator1* and *Manipulator2*, acting as foundations for their child objects *Detal1, Detal2, Detal3* and *Detal11, Detal22, Detal33*. Script files *Manyapul1.cs* and *Manyapul2.cs* were created specifically for these objects. Afterwards we have created game objects for the chess pieces that contain script file *FigureScript.cs* and which are child game objects of *WhitePlayer* and *BlackPlayer*.

**Algorithm and Program Implementation.** Main program work is set and controlled by script *BattleField.cs* that receives requests to make the move and to use the value of variables that contain information about the move. Depending on who is going to make the move (which object sent the request), the script provides the manipulator robots MR1 and MR2 with values of variables and the request to make a move.

Depending on the current code, the scripts *Manyapul1.cs* and *Manyapul2.cs* receive the request to make a move together with current values of variables. Based on the values of variables, the scripts provide the manipulator's elements with the values which shall be used to open its tiers in order to move the grip to the required square. Then the script launches the turn function *PovorotFrom*, which is responsible for the manipulator's turn, further informing the chess piece's script about its move and after that launching the function *PovorotTo*.

*PovorotTo* function provides manipulator's elements with the values, according to which they should be turned in order to move the grip towards the target square, then sending a moving message to the chess piece's script file and putting the manipulator into motion. Afterwards the program delays the manipulator's movement for the time that can be set by the vertical sliders located on the screen; then *DeFolt* function is launched. *DeFolt* function provides the manipulator elements with the source location values, performing the manipulator's turn and sending the move-end message to the script file *Main.cs*.

Program scripts contain several service functions, whose values we are not going to describe here due to a limited size of the publication.

After launching the program we can see a configuration window, where we can make changes in the current settings of graphic interface and control elements. The settings are saved in the configuration file. The game window is opened after pressing Play button (see Pic. 4).

In the current version of the simulation program we can only replay the sequence of moves that were recorded previously in the file *Game.txt*. Should the user wish to replay his/her own game, he/she should create his/her own file *Game.txt* and place it into the folder that contains the executable file *chessgame.exe*.

In the game window the user can access the following control elements together with the data display fields (see Pic. 4):

• The slider that changes movement speed of the black manipulator (see red mark 1); the higher the slider is, the faster the manipulator moves (black - mark 8, white - mark 9).

• The slider that changes movement speed of the white manipulator (mark 3); the higher the slider is, the faster the manipulator moves.

• *Exit* button (mark 5), pressing which the user stops the application.

• POSHAGOVO? (STEP BY STEP?) button (mark 6) launches the game in a step-by-step mode, i.e. the manipulators shall stop after each step.

• *Autoplay?* button (mark 7) launches the game in an automatic mode, where the manipulators shall make moves one after another until the game is over.

• Khodim? (Go?) button (see mark 1 on Pic. 5) appears only when the step-by-step mode has been selected; if the move has already been made, the next move shall be made only when pressing the button.

- Nachnem? (Start?) button appears after the launch game option was chosen, the first move is made after pressing it.
- Text fields "Black Manipulator's Move Time: 0.8" (mark 2 on Pic. 5) and "White Manipulator's Move Time: 1.2" (mark 3 on Pic. 5) display duration of the last move, made accordingly by black and white MRs.
- In the beginning of the game the chess pieces are in their source locations, i.e. the black are on the left (mark 10) and the white are on the right (mark 11).

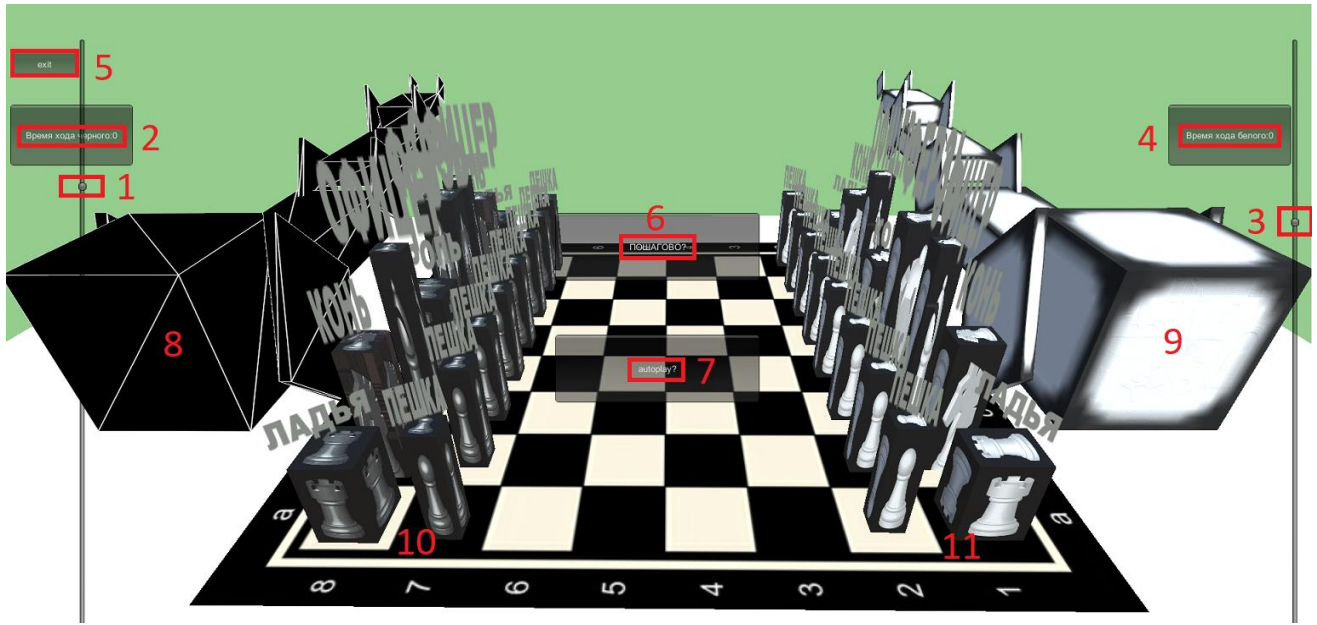

Picture 4 - Game Window. Source location of chess pieces and control elements

Picture 5 shows game position after the manipulators have made several moves. The manipulator with white chess pieces (mark 4) is completing its move.

The program provides accumulation and output of measurable data per each simulated game. In particular, it is possible to measure the length of the grip trajectory as well as the time required for each move.

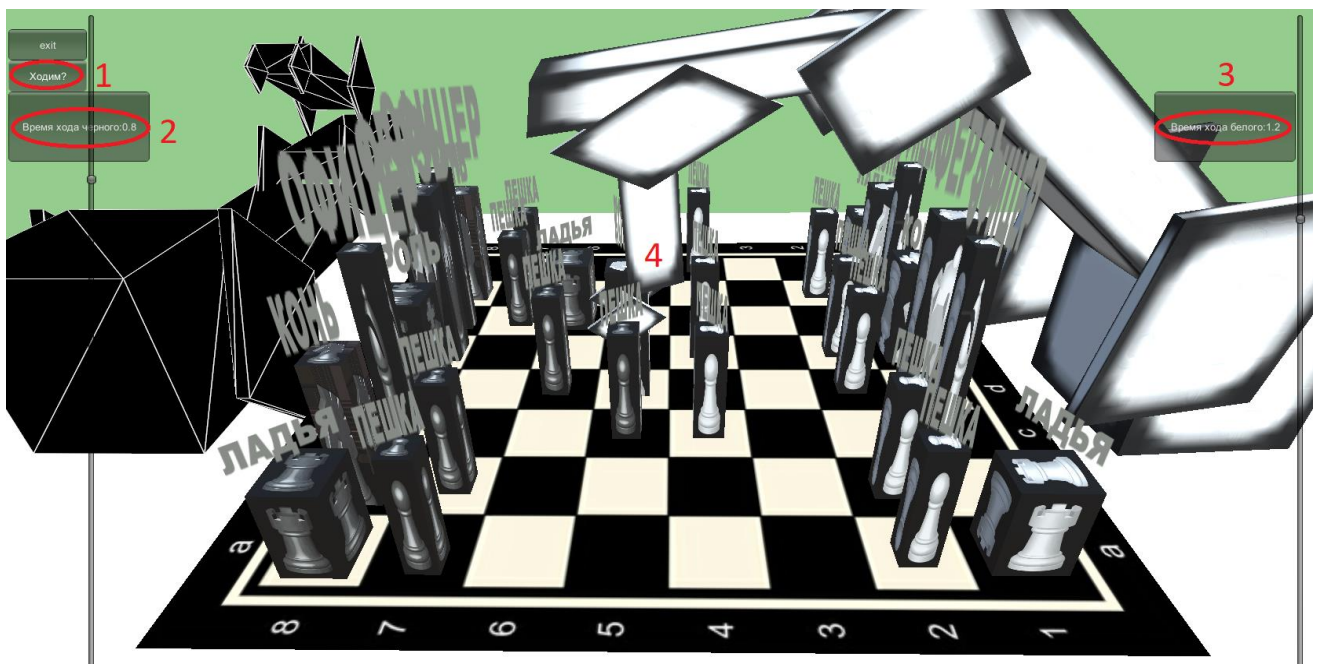

Picture 5 - Intermediate Position of the Game

These indicators can be used to optimize the movement of manipulators, for example in relation to the operating speed and/or power consumption.

Picture 6 shows registered duration of each move made by the manipulators.

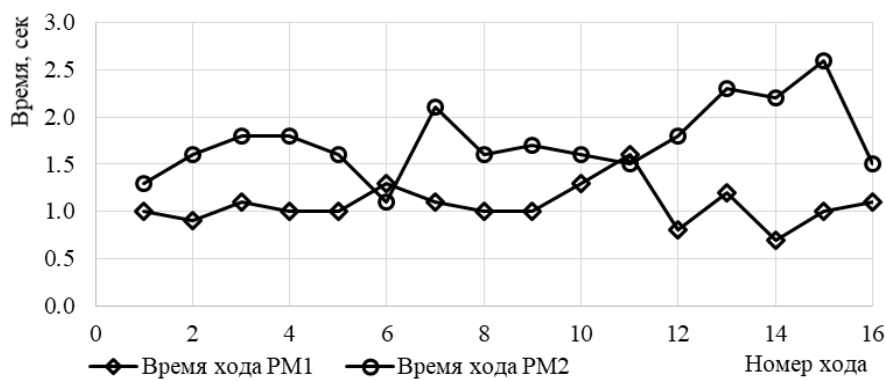

Picture 6 - Move Duration Per Each Manipulator

Statistical characteristics of this data are the following: average value of MR1 making its move is 1.07 sec, MR2 - 1.76 sec, while the correlation coefficient value of these two datasets constitutes $r_{12} = -033$. These three indicators can be used when solving the problem of manipulator optimization, i.e. tier size, location of their bases, choosing the type of a servo drill etc.. It is also necessary to mention that this model can be launched in a virtual mode (without screen image but with measurement of all parameters) thus giving a possibility to apply optimization search algorithms.

**Results and Discussion.** Provided simulation game model is only a part of the simulation & nature robotic complex, which shall be used for further research, working over the construction elements as well as the work algorithms.

Nowadays, the work of autonomous manipulator robots or android robots in a non-determined environment is a highly sought after and necessary topic [1, 3, 4]. It can be applied at the space station when assembling its construction elements, at other planets or on the ground when building various objects etc.. At that, the operating range of MR can vary greatly: from the static set of homogeneous objects to a large variety of diverse dynamic objects [4]. Here, the variety of possible operating environments is also determined by the fact that MRs can be both stationary or moving, located in a stable and spatially-linked group or as a totality of independently moving MRs.

The chess context allows working over structure and combination of algorithmic and program elements, their cooperation as well as optimizing the parameters.

At the next stage we expect to add a free chess engine to the current simulation model [14]. It is obvious that a full-function simulation model of MR playing chess does not allow working over some situations that may occur in real conditions of the above-mentioned spheres. Thus, the model does not include sensory capabilities which are required when manipulating the objects in a real time. Therefore, the natural part of the model complex shall include stage surveillance cameras, corresponding situation (condition, scene) identification algorithms, as well as other means of controlling the work area.

Apart from that, the environment of the observed simulation & nature complex shall include various grip options for various objects (chess pieces): two- and three-finger grips (also the collet type) as well as a five-finger anthropomorphic hand [15-18]. It would also be interesting to perform a research on the grip construction, including the feedback on strengthening the grip with installed strain indicators and actuators, installed in the human's operating gear.

Apart from the autonomous work of the group of (now only two) MRs, there should also exist a possibility to implement a human-machine mode where one MR is to be controlled by the human operator using various interface options, including somatosensory glove (Exoskeleton), gyroscope sensors, neurobionical helmet and neurosensors on the hand. Here, the computing environment shall include elements of Arduino family [19] (connected to the personal computer) and the robot development environment EZ-Robot [20]. These options allow implementing control algorithms practically of any complexity.

Human-machine mode would also give a possibility of working over some collaborative work modes of MRs [3].

The simulation environment shall provide a possibility of working on adaptive training algorithms where MR would perform actions based on observation (registration) of the human operator's actions, for instance using a somatosensory glove [19] or video cameras.

The above-mentioned functional capabilities of the created simulation & nature robotic complex are mostly focused on the training objectives applied from college level to the doctoral studies. However, its modular nature together with the possibility to replace parts of real elements with the simulation modules allows performing research that would also prove useful to the application in practice.

**Conclusions**

1. The developed totality of the chess game simulation modelling proved the possibility of building the cooperation model for manipulator robots in a chess game. Visual demonstration of the game course shows the cooperation of the manipulators, while the control elements provide a possibility of showing the results as it is desired, when changing some modelling parameters.

2. The modelling program is built in such a way that it allows performing different kinds of research of the manipulator robot cooperation, in particular, changing time characteristics of each MR's actions.